\documentclass[twocolumn,10pt,journal,twoside]{IEEEtran}
\usepackage[cmex10]{amsmath}
\usepackage{amssymb}
\usepackage{amsfonts}
\usepackage{amsthm}
\usepackage{algorithm}
\usepackage{algpseudocode}
\usepackage{graphicx}
\usepackage{epsfig}
\usepackage{cite}
\usepackage{tensor}
\usepackage[caption=false,font=footnotesize]{subfig}
\usepackage{color}
\usepackage{makecell}
\usepackage{float}
\usepackage{comment}
\usepackage{subfig}
\usepackage[dvipsnames]{xcolor}
\usepackage{algpseudocode}

\usepackage{hyperref}
\hypersetup{hypertex=true,
	colorlinks=true,
	linkcolor=blue,
	anchorcolor=blue,
	citecolor=blue}

\graphicspath{{figures/}}
\DeclareGraphicsExtensions{.eps}
\theoremstyle{definition}
\newtheorem{remark}{Remark}

\newcommand\T{{\hspace{-0pt}\intercal}}

\DeclareMathOperator{\diag}{diag}

\DeclareMathOperator{\trace}{trace}

\begin{document}

\title{
Adaptive Shape-Servoing for Vision-based Robotic Manipulation with Model Estimation and Performance Regulation
}

\author{
Fangqing Chen $^*$\\
University of Toronto

\thanks{
Copyright may be transferred without notice, after which this version
may no longer be accessible.
}
\thanks{
$^*$ Corresponding Author.
}

}

\bstctlcite{IEEEexample:BSTcontrol}

\maketitle

\begin{abstract}
This paper introduces a manipulation framework for the elastic rod, including shape representation, sensorimotor-model estimation, and shape controller.
Until now, the manipulation of the elastic rod has faced several challenges:
1) shape learning from high-dimensional to low-space dimensional;
2) the modeling of robot manipulation of the elastic rod;
3) the determination of the shape controller.
A novel manipulation framework for the elastic rod is presented in this paper, which only uses the input and output data of the system without any prior knowledge of the robot, camera, and object.
The proposed approach runs in a model-free manner.
For the approximation of the sensorimotor model, adaptive Kalman filtering (AKF) is used as the online estimation.
Model-free adaptive control (MFAC) is designed according to the obtained differential model of robot-object configuration and then is combined with the performance regulation requirement to give the final format of the shape controller.
Hence, the proposed approach can enhance the autonomous capability of deformation object manipulation.
Detailed simulation results are conducted with a single robot manipulation to evaluate the effectiveness of the proposed manipulation framework.

\end{abstract}

\IEEEpeerreviewmaketitle

\section{Introduction}\label{section1}
Shape control of the deformable object with robot manipulation automatically is highly valuable in many applications, such as 
industrial production \cite{taipale2015robot},
wound suturing \cite{sagitov2018automated}, 
item packaging \cite{balatti2021flexible}, and 
household service \cite{yang2023integrating}. 
Although great achievements have been obtained in the field of robotic manipulation in recent years, shape control of deformable objects is still a hot issue \cite{arriola2020modeling}. 
One of the biggest technological difficulties of implementing these types of controls is the difficulty of modeling the differential relationship between the robot and the manipulated object during the manipulation process.

The rigid object's configuration (refer to pose state here) can be easily described by six degrees of freedom (DOF) \cite{schmidt2016perception}.
However, different from rigid objects, deformable objects are hard to be compatible with simple 6DOF (position plus pose) for their shape representation \cite{yang2019evaluating}.
Meanwhile, since deformable objects have complex and unknown kinematic models, achieving ideal control effects with traditional model-based control methods is difficult.
Especially in practice, their corresponding control performance is often greatly compromised when it comes to deformable objects of different types and sizes \cite{strecke2021diffsdfsim}.

To the best of our knowledge, this is the first attempt to design a model-free adaptive controller and simultaneously consider performance regulation for the deformation object manipulation.
The detailed simulations are conducted on the single-robot setup with the eye-to-hand configuration, and our model is compared against several state-of-the-art methods. 
The results demonstrate that the proposed framework not only outperforms the existing methods in terms of deformation error but also exhibits greater convergence speed near the neighborhood of the origin, which is more important in real industrial production.
Moreover, the proposed approach gives a manipulation framework of the elastic rod from the control viewpoint, offering shape presentation, differential relationship approximation, and shape controller design underlying criminal performance regulation requirements.

The key contributions of this paper are three-fold:
\begin{itemize}
\item 
\textbf{Construction of the Deformable Object Manipulation:}
This paper presents the manipulation framework of the elastic rod from the control's viewpoint, including representation, estimation, and shape control.
This novel approach does not need prior knowledge of the system's information, e.g., robot, camera, and manipulated object, fully running model-free.

\item 
\textbf{Development of the Online Differential Model Estimator:}
Adaptive Kalman Filtering is used to estimate the sensorimotor model between the robot and the elastic rod.
It relies only on input and output data and has fewer tuning parameters than existing approaches.
This paper constructs a framework to estimate the differential model in the format of the augmented system.

\item 
\textbf{Development of the MFAC with Performance Regulation}
This paper innovatively proposes a model-free adaptive control considering performance tuning for shape control.
It not only considers the absolute error of shape deformation but also addresses the change rate of deformation error and the performance of the robot control instructions (i.e., position, velocity, acceleration),

\end{itemize}

\section{Related Works}
In this section, we review the existing literature relevant to our research.
This discussion is divided into two aspects, i.e., shape representation and performance regulation.

\subsection{Shape Representation}
For manipulation, the first step is to construct a low-dimensional feature that can learn the main shape characteristics \cite{zuffi20173d,smith1995shape, qureshi2007graph}.
Traditional representation methods are roughly divided into local and global descriptors. 
To describe geometric characteristics, local descriptors may use
central \cite{mack2017experimental}, 
distances \cite{zhao2014modulating}, 
angles \cite{el2008novel}
, and curvatures \cite{worring1993digital}. 
However, these features can only work for the local range and are not good for the case of a wide range of deformation.
In contrast, global descriptors produce an overall representation of the object.
One example is that the Point Feature Histogram (PFH) reported in \cite{jin2013new} constructs a low-dimensional feature (low to 33) to represent the overall sponge's shape, then is improved into the Fast Point Feature Histograms (FPFH) \cite{rusu2009fast,zhao2016point}.
However, the above techniques usually offer a high-dimensional feature vector, which may cause the calculation burden of the system or even the manipulation failure.

\subsection{Sensorimotor Model Approximation}
To visually guide the manipulation task, the system must know the format of the kinematic model (at least within the approximation state), which demonstrates how the robot's motion produces the shape's changes \cite{ma2022learning}.
In the field of visual servoing, such differential relationship is usually named after the deformation Jacobian matrix (DJM) \cite{weiss1985dynamic,jin2019long,hutchinson1996tutorial}.
Much research have been conducted to solve this estimation, e.g., the
Broyden iteration method \cite{hosoda1994versatile} is used to estimate this dynamic transformation model \cite{alambeigi2018autonomous,lagneau2020automatic,kawata1998efficient}. 
Although these types of algorithms do not require knowledge of the model's structure, their estimation properties are only valid locally \cite{aghajanzadeh2022offline}.
Most existing methods estimate the robot-object model considering a single performance criterion \cite{angrist1995identification} (typically, the absolute estimation accuracy), which does not pay attention to the dynamic responses during the manipulation process.

\section{PROBLEM FORMULATION}\label{section2}
\emph{Notation.}
Throughout this paper, we use very standard notation.
Column vectors are denoted with bold small letters $\mathbf{v}$, and matrices with bold capital letters $\mathbf{A}$.
Time evolving variables are represented as $\mathbf{m}_k$, where the subscript $k$ denotes the discrete time instant.
The all-ones matrix is denoted by $\mathbf{I}_{m \times n}$, and $\mathbf{E}_n$ is an $n \times n$ identity matrix.
$\otimes$ represents the Kronecker product.

\subsection{Robot Model}\label{section2a}
Let us consider the configuration of a single robot, and the vectors of joint angle and the effector's pose are defined by $\mathbf{q} \in \mathbb{R}^d$ and $\mathbf{r} \in \mathbb{R}^{q}$, respectively.
The differential kinematics of the single are introduced as follows:
\begin{equation}
\dot{\mathbf{r}} = \mathbf{J}_r (\mathbf{q}) \dot{\mathbf{q}}
\end{equation}
where $\mathbf{J}_r(\mathbf{q})$ is called by the robot Jacobian matrix (RJM); it describes the velocity relationship between the joint velocity of the robot and the change rate of the effector's pose.
In common, $\mathbf{J}_r(\mathbf{q})$ is assumed to be known in advance.
In this article, we consider the robot to be kinematically controlled (i.e., the robot can be exactly controlled without time delay), and without collision with obstacles or the robot itself.

\subsection{Shape Perception Model}
In this work, we consider a vision-guided robot manipulation system.
A camera has an eye-to-hand configuration, and the effector is connected with the manipulated object in advance and then starts to deform the object to meet the desired requirements.
The 2D centerline shape of the object is described as:
\begin{equation}
\label{eq41}
{\bar{\mathbf{c}} = {{\left[ {\mathbf{c}_1^\T, \dots ,\mathbf{c}_N^\T} \right]}^\T} \in {\mathbb R^{2N}}}, \ {{\mathbf{c}_i} = {{\left[ {{c_{ui}},{c_{vi}}} \right]}^\T} \in {\mathbb R^2}}
\end{equation}
where $N$ is the number of the centerline points, $c_{ui}$ and $c_{vi}$ are the pixel coordinates of the $i$th image point.

\subsection{Deformation-Motion Sensorimotor Model}\label{section2b}
Given that regular elastic rod are considered in this work, the centerline $\bar{\mathbf{c}}$ is strictly dependent on the effector's pose $\mathbf{r}$.
The kinematic relationship between $\bar{\mathbf{c}}$ and $\mathbf{r}$ is modeled as:
\begin{equation}
\label{eq42}
    \bar{\mathbf{c}} = \mathbf{f}_c (\mathbf{r})
\end{equation}

For subsequent the deformation task, a reduced-dimension feature $\mathbf{s} \in \mathbb{R}^p$ is designed to represent the full centerline shape of the object in the low-dimensional space, where $p \ll 2N$.
The main goal of the shape feature $\mathbf{s}$ is to avoid system instability caused by excessively large original data dimensions.
In principle, $\mathbf{s}$ and $\bar{\mathbf{c}}$ have a one-to-one correspondence with the following expression:
\begin{equation}
\label{eq43}
    \mathbf{s} = \mathbf{f}_s (\bar{\mathbf{c}})
\end{equation}

Combined with \eqref{eq42} and  \eqref{eq43}, it has
\begin{equation}
\label{eq44}
    \mathbf{s} = \mathbf{f}_s ( \mathbf{f}_c (\mathbf{r}))
\end{equation}

Calculating the derivative of \eqref{eq44} with respect to time, the first-order kinematic model is obtained:
\begin{equation}
\label{eq46}
    \dot{\mathbf{s}} = \frac{\partial \mathbf{f}_s}{ \partial \mathbf{r}} \dot{\mathbf{r}}
     = \mathbf{J}_s(\mathbf{r}) \dot{\mathbf{r}}
\end{equation}
where $\mathbf{J}_s(\mathbf{r})$ is the deformation Jacobian matrix (DJM).
As the properties of elastic rods are unknown, the analytical form of $\mathbf{J}_s(\mathbf{r})$ cannot be obtained. 
Discretizing \eqref{eq46} yields the first-order format as follows:
\begin{equation}
\label{eq47}
\mathbf{s}_k = \mathbf{s}_{k-1} + \mathbf{J}_k \Delta \mathbf{r}_k
\end{equation}
where $\Delta \mathbf{r}_k = \mathbf{r}_k - \mathbf{r}_{k-1} \in \mathbb{R}^q$ is the velocity command of the effector's pose.

\begin{remark}
In the following content, if there are no special instructions, we omit variables in the bracket for the convenience of writing, e.g., $\mathbf{J}_s(\mathbf{r})$ is replaced by $\mathbf{J}_s$.
\end{remark}

\section{AKF-based for DJM Approximation}
As this work's point is to propose a manipulation framework for the elastic objects, the simple adaptive Kalman filtering (AKF) is used to compute the local approximations of $\mathbf{J}_s$.
Two augmented variables are defined by:
\begin{equation}
\label{eq1}
\begin{array}{*{20}{c}}
{\Delta {\mathbf{s}_k} = {\mathbf{s}_k} - {\mathbf{s}_{k - 1}} \in \mathbb{R}^p},&
{\Delta {\mathbf{r}_k} = {\mathbf{r}_k} - {\mathbf{r}_{k - 1}} \in \mathbb{R}^q}
\end{array}
\end{equation}

We define ${\mathbf{y}_k = \Delta \mathbf{s}_k}$ and ${\mathbf{u}_k = \Delta \mathbf{r}_k}$.
With \eqref{eq47} yields
\begin{equation}
\label{eq3}
\mathbf{s}_k = {\mathbf{J}_k}{\mathbf{u}_k}
\end{equation}

Define the following augmented variables:
\begin{align}
\label{eq25}
{\mathbf{x}_k} &= {\left[ {\frac{{\partial {s_1}}}{{\partial {\mathbf{r}^\T}}}, \ldots ,\frac{{\partial {s_p}}}{{\partial {\mathbf{r}^\T}}}} \right]^\T} \in {\mathbb{R}^{pq}} \notag\\
\frac{{\partial {s_i}}}{{\partial {\mathbf{r}^\T}}} &= \left[ {\frac{{\partial {s_i}}}{{\partial {r_1}}}, \ldots ,\frac{{\partial {s_i}}}{{\partial {r_q}}}} \right] \in {\mathbb{R}^{1 \times q}}
\end{align}

Combing \eqref{eq3} and \eqref{eq25}, we have the state-space model:
\begin{align}
\label{eq4}
{\mathbf{x}_k} = {\mathbf{A}_k}{\mathbf{x}_{k - 1}} + {\eta _k}, \ \ \
{\mathbf{y}_k} = {\mathbf{M}_k}{\mathbf{x}_k} + {\nu _k}
\end{align}
where $\mathbf{A}_k$ is the transition matrix, which is a unit matrix in principle.
$\mathbf{M}_k$ is the observation matrix with the following definition:
\begin{equation}
\label{eq5}
\mathbf{M}_k = \diag \left(
\begin{matrix} \underbrace{ \Delta \mathbf{r}_k^\T,\cdots, \Delta \mathbf{r}_k^\T } \\ p
\end{matrix} 
\right) \in {\mathbb R^{p \times pq}}
\end{equation}

For the traditional Kalman filtering (KF), it is usually supposed that the state model noise and the measurement noise are both known Gaussian whiter noise for the optimal estimation performance in the practical system.
However, the probability of noise usually varies during the estimation period, and the noise matrices are not simple Gaussian white noise most of the time.
To solve this issue, AKF is proposed as a remedy to this problem, which is explained in the sequel.
The calculation process of AKF is given by:
\begin{align}
\label{eq28}
{\mathbf{K}_k} 
&= \frac{1}{{{\alpha _k}}}{\mathbf{P}_{k|k - 1}} \mathbf{C}_k^\T{\left( {\frac{1}{{{\alpha _k}}}{\mathbf{C}_k}{\mathbf{P}_{k|k - 1}}\mathbf{C}_k^\T 
+ {{\hat{\mathbf{R}} }_k}} \right)^{ - 1}} \notag \\
{{\hat{\mathbf{x}} }_k}
&= {{\hat{\mathbf{x}} }_{k - 1}} + {\mathbf{K}_k}\left( {{\mathbf{z}_k} - {\mathbf{C}_k}{{\hat{\mathbf{x}} }_{k - 1}}} \right) \notag \\
{\mathbf{P}_{k|k}} 
&= \frac{1}{{{\alpha _k}}}\left( {{\mathbf{E}_{pq}} - {\mathbf{K}_k}{\mathbf{C}_k}} \right){\mathbf{P}_{k|k - 1}}
\end{align}
where $\alpha_k$ is an adaptive factor that is calculated by
\begin{equation}
\label{eq29}
{\alpha _k} = \left\{ {\begin{array}{*{20}{c}}
	1&{\left| {\Delta {\varepsilon _k}} \right| \le {c_0}}\\
	{\frac{{{c_0}}}{{\left| {\Delta {\varepsilon _k}} \right|}}{{\left( {\frac{{{c_0} - \left| {\Delta {\varepsilon _k}} \right|}}{{{c_1} - {c_0}}}} \right)}^2}}&{{c_0} < \left| {\Delta {\varepsilon _k}} \right| \le {c_1}}\\
	0&{\left| {\Delta {\varepsilon _k}} \right| > {c_1}}
	\end{array}} \right.
\end{equation}
where $\Delta \varepsilon_k$ is a learning statistic constructed by predicted residual vector.
$c_0$ and $c_1$ are two positive constants, with practical values, $c_0 \in [1,1.5]$ and $c_1 \in [3,8]$.
Then, $\Delta \varepsilon_k$ can be defined as:
\begin{equation}
\label{eq30}
\Delta {\varepsilon _k} = \sqrt {\frac{{\varepsilon _k^T{\varepsilon _k}}}{{\trace \left( {{\mathbf{G}_k}} \right)}}} 
\end{equation}
where $\varepsilon_k$ and $G_k$ are the predicted residual vector and the theoretical covariance matrix of the predicted residual vector, respectively.
\begin{align}
\label{eq31}
\begin{array}{*{20}{c}}
{{\varepsilon _k} 
= {\mathbf{z}_k} - {\mathbf{C}_k}{{\hat{\mathbf{x}} }_{k|k - 1}}}, \ \ \
{{\mathbf{G}_k} = {\mathbf{C}_k}{\mathbf{P}_{k|k - 1}} \mathbf{C}_k^\T}
\end{array}
\end{align}

The state model and measurement of noise covariance matrices can be written in the form of a recursive estimation formula:
\begin{align}
\label{eq32}
{{\hat{\mathbf{R}} }_k} 
&= \left( {1 - {d_k}} \right){{\hat{\mathbf{R}} }_{k - 1}} + {d_k}\left( {{\varepsilon _k}\varepsilon _k^\T - {\mathbf{C}_k}{\mathbf{P}_{k|k - 1}} \mathbf{C}_k^\T} \right)\\
{{\hat{\mathbf{Q}} }_k}
&= \left( {1 - {d_k}} \right){{\hat{\mathbf{Q}} }_{k - 1}} + {d_k}\left( {{\mathbf{K}_k}{\varepsilon _k}\varepsilon _k^\T \mathbf{K}_k^\T + {\mathbf{P}_{k|k}} - {\mathbf{P}_{k|k - 1}}} \right) \notag
\end{align}
where ${d_k} = \left( {1 - b} \right)/\left( {1 - {b^{k + 1}}} \right)$ is a correction factor, and $b \in (0,1)$ is a forgotten factor which specifies the weight of the history data.
In practice, $\hat{\mathbf{R}}_k$ may gradually lose the positive-definite value.
When the filter gain is more than 1, the filter is prone to present a divergent state.
Thus, $\hat{\mathbf{R}}_k$ could be improved as follows:
\begin{equation}
\label{eq33}
{\hat{\mathbf{R}} _k} = \left( {1 - {d_k}} \right){\hat{\mathbf{R}} _{k - 1}} + {d_k}{\varepsilon _k}\varepsilon _k^\T
\end{equation}

Notice that, the estimator of $\mathbf{R}_k$ is calculated at the expense of the unbiasedness; it is likely to improve the stability.
Once we get $\hat{\mathbf{x}}_k$ at each step, we update the DJM online according to \eqref{eq25}.
From the above analysis, it can be seen that the estimation of DJM only uses input and output data, and does not require the physical parameters of the robot and the object itself. 
Therefore, this is an online adaptive approximation measure.

\section{MFAC-based Velocity Controller}
At the discrete-time instant $k$, we suppose that $\mathbf{J}_k$ has been by estimated by $\hat{\mathbf{J}}_k$ accurately, so that the robot-object differential model is satisfied:
\begin{equation}
\label{eq19}
{\mathbf{s}_k} = {\mathbf{s}_{k - 1}} + {\hat{\mathbf{J}}_k}{\mathbf{u}_k}
\end{equation}

Define the deformation error, $\mathbf{e}_k = \mathbf{s}^* - \mathbf{s}_k$, where $\mathbf{s}^*$ is the constant target feature representing the desired centerline shape.
The novel objective function is defined as follows:
\begin{align}
\label{eq20}
Q\left( {{\mathbf{u}_k}} \right) 
&= {\omega _1}{\left\| {{\mathbf{s}^*} - {\mathbf{s}_k}} \right\|^2} 
+ {\omega _2}{\left\| {{\mathbf{s}_k}} \right\|^2} \notag \\
&+ {\omega _3}{\left\| {{{{\mathbf{s}_k} - {\mathbf{s}_{k - 1}}}}} \right\|^2} 
+ {\omega _4}{\left\| {{{\Delta {\mathbf{s}_k} - \Delta {\mathbf{s}_{k - 1}}}}} \right\|^2}\\
&+ {\omega _5}{\left\| {{\mathbf{r}_k}} \right\|^2} 
+ {\omega _6}{\left\| {{\mathbf{r}_k} - {\mathbf{r}_{k - 1}}} \right\|^2} 
+ {\omega _7}{\left\| {{\mathbf{u}_k} - {\mathbf{u}_{k - 1}}} \right\|^2} \notag
\end{align}
where $\sum\limits_{i = 1}^7 {{\omega _i}}  = 1$ are positive constant values, which are used to adjust the weight of each criterion item.
And, we have:
\begin{align}
\label{eq21}
{\mathbf{s}^* - \mathbf{s}_k} &= {\mathbf{e}_{k - 1}} - {{\hat{\mathbf{J}}}_k}{\mathbf{u}_k} \notag \\
\Delta {\mathbf{s}_k} - \Delta {\mathbf{s}_{k - 1}} &= {{\hat{\mathbf{J}}}_k}{\mathbf{u}_k} - {{\hat{\mathbf{J}}}_{k - 1}}{\mathbf{u}_{k - 1}}
\end{align}

The role of each item is described below:
\begin{itemize}
	\item $\omega_1$ limits the amplitude of $\mathbf{e}$;
	
	\item $\omega_2$ limits the amplitude of $\mathbf{s}$;
	
	\item $\omega_3$ limits the amplitude of $\Delta \mathbf{s}$;
	
	\item $\omega_4$ limits the increment of $\Delta \mathbf{s}$;
	
	\item $\omega_5$ limits the amplitude of $\mathbf{r}$ to prevent the robot out from the workspace;
	
	\item $\omega_6$ limits the amplitude of $\mathbf{u}$ to prevent excessive displacement of the robot;
	
	\item $\omega_7$ limits the increment of $\mathbf{u}$ (i.e., the acceleration) to ensure the continuity of $\mathbf{u}$;
	
\end{itemize}

By substituting \eqref{eq21} into \eqref{eq20}, it yields
\begin{align}
\label{eq22}
Q\left( {{\mathbf{u}_k}} \right) 
&= {\omega _1}{\left\| {{\mathbf{e}_{k - 1}} - {{\hat{\mathbf{J}} }_k}{\mathbf{u}_k}} \right\|^2} 
+ {\omega _2}{\left\| {{\mathbf{s}_{k - 1}} + {{\hat{\mathbf{J}} }_k}{\mathbf{u}_k}} \right\|^2} \notag \\
&+ {\omega _3}{\left\| {{{\hat{\mathbf{J}} }_k}{\mathbf{u}_k}} \right\|^2} 
+ {\omega _4}{\left\| {{{\hat{\mathbf{J}} }_k}{\mathbf{u}_k} - {{\hat{\mathbf{J}} }_{k - 1}}{\mathbf{u}_{k - 1}}} \right\|^2}\\
&+ {\omega _5}{\left\| {{\mathbf{r}_{k - 1}} + {\mathbf{u}_k}} \right\|^2} 
+ {\omega _6}{\left\| {{\mathbf{u}_k}} \right\|^2} +
{\omega _7}{\left\| {{\mathbf{u}_k} - {\mathbf{u}_{k - 1}}} \right\|^2} \notag 
\end{align}

Calculate the partial derivative of \eqref{eq22} w.r.t. $u_k$, it yields
\begin{align}
\label{eq23}
\frac{{\partial Q\left( {{\mathbf{u}_k}} \right)}}{{\partial {\mathbf{u}_k}}} 
&= 2\left( \begin{array}{l}
 - {\omega _1}\hat{\mathbf{J}} _k^\T{\mathbf{e}_{k - 1}} 
+ {\omega _2}\hat{\mathbf{J}} _k^\T{\mathbf{s}_{k - 1}} 
+ {\omega _5}{\mathbf{r}_{k - 1}}\\
- \left( {{{{\omega _4}}}\hat{\mathbf{J}} _k^\T{{\hat{\mathbf{J}} }_{k - 1}} 
	+ {\omega _7}\mathbf{I}_q} \right){\mathbf{u}_{k - 1}}
\end{array} \right) \notag \\
&+ 2\left( \begin{array}{l}
\left( { \omega _1 + \omega _2 + \omega _3 + \omega _4} \right)
\hat{\mathbf{J}} _k^\T{{\hat{\mathbf{J}} }_k}\\
+ \left( {{\omega _5} + {\omega _6} + {\omega _7}} \right){\mathbf{I}_q}
\end{array} \right){\mathbf{u}_k}
\end{align}

By equating \eqref{eq23} to zero, the velocity controller $\mathbf{u}_k$ can be calculated as
\begin{align}
\label{eq24}
\mathbf{u}_k &= \mathbf{G}_k
\left( 
\begin{array}{l}
{\omega _1}\hat{\mathbf{J}} _k^\T{\mathbf{e}_{k - 1}} 
- {\omega _2}\hat{\mathbf{J}} _k^\T{\mathbf{s}_{k - 1}} 
- {\omega _5}{\mathbf{r}_{k - 1}} \notag \\
+ \left( {{{{\omega _4}}}\hat{\mathbf{J}} _k^\T{{\hat{\mathbf{J}} }_{k - 1}} 
+ {\omega _7}\mathbf{I}_q} \right){\mathbf{u}_{k - 1}}
\end{array} 
\right)\\
\mathbf{G}_k &= 
\left( 
\begin{array}{l}
\left( {\omega _1} + {\omega _2} + \omega _3 + \omega _4 \right)
\hat{J} _k^\T{{\hat{\mathbf{J}} }_k}\\
+ \left( 
{{\omega _5} 
+ {\omega _6} 
+ {\omega _7}} 
\right){\mathbf{I}_q}
\end{array}
\right)^{ - 1}
\end{align}

The satisfactory control performance can be obtained by selecting appropriate parameters $\omega_i, i\in [1,7]$.

\section{Experiment Results}
In this section, the experimental setup and the results of the proposed manipulation framework are conducted.

\subsection{Extraction of the  Shape Feature}
For the sake of simplicity of the article, we directly adopt the feature extraction algorithm presented in \cite{qi2022towards}, i.e., the Auto-encoder feature extraction network, and we set the number of the latent layer is $n=6$, which means the dimension of the feature used in this work is $p = 6$.

For the neural network training, 100,000 samples $(N=100)$ are collected by controlling the robot to deform the object continuously.
The Auto-encoder network is built on PyTorch with ADAM optimizer with an initial learning rate of 0.001, and the batch size is set to 10,000.
RELU function is adopted in the Encoder and Decoder sections, respectively.

\subsection{Evaluation of the Manipulation}
In this section, we evaluate the effectiveness of the proposed manipulation framework by commanding the robot to deform the elastic rod into the desired configuration.
For a fair comparison, two frameworks \cite{ma2022active} and \cite{aghajanzadeh2022optimal} are conducted, including representation, estimation and shape control in a full manipulation manner.
The deformation metric is defined as:
\begin{equation}
T_1 = \| \mathbf{s}^* - \mathbf{s}_k  \|
\end{equation}

\begin{figure}[htbp]
\centering
\includegraphics[scale=0.062]{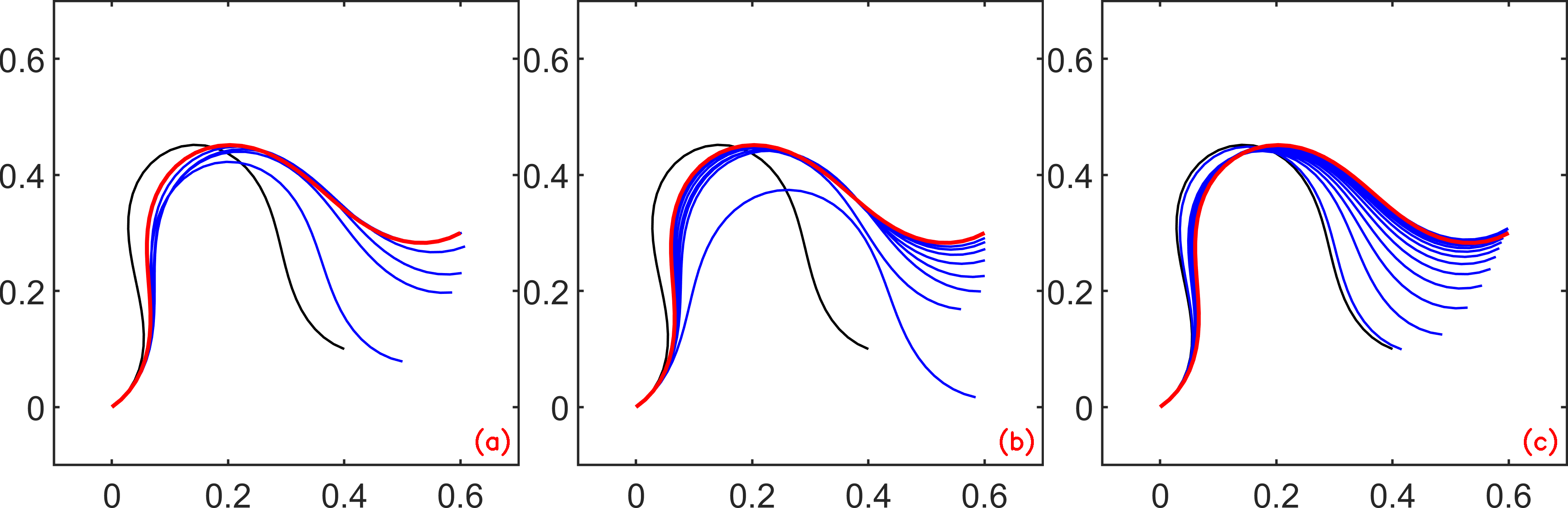}
\caption{
Profiles of the shape deformation simulation among 
Ours (a), 
Ma et al. \cite{ma2022active} (b), and 
Omid et al. \cite{aghajanzadeh2022optimal} (c).
The black solid curve represents the initial shape, the blue curves give transitional trajectories, and the red solid curve describes the desired shape $\bar{\mathbf{c}}^*$, with response to the desired feature $\mathbf{s}^*$.
The deformation trajectories display every three frames for the concrete comparison.
The abscissa is the step size.
}
\label{fig3}
\end{figure}

\begin{figure}[htbp]
\centering
\includegraphics[scale=0.12]{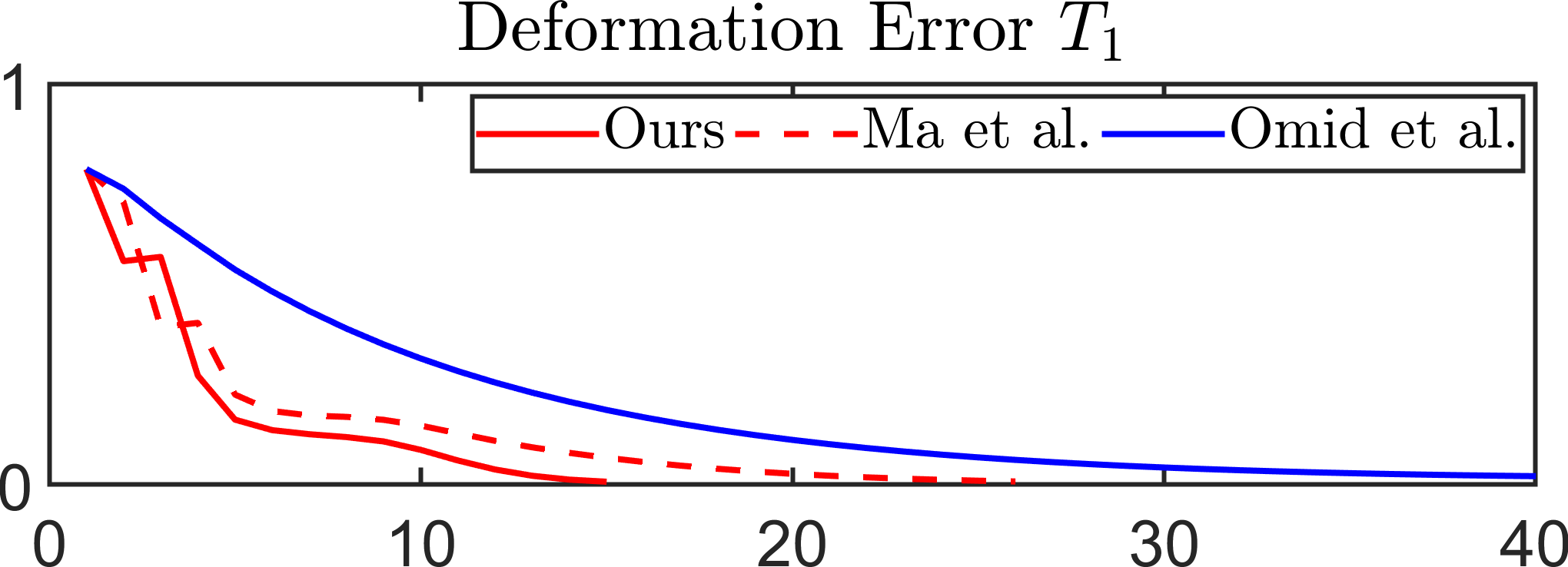}
\caption{
Profiles of the deformation error $T_1$ among Ours, Ma et al. \cite{ma2022active}, and Omid et al. \cite{aghajanzadeh2022optimal}.
The abscissa is the step size.
}
\label{fig1}
\end{figure}

\begin{figure}[htbp]
\centering
\includegraphics[scale=0.11]{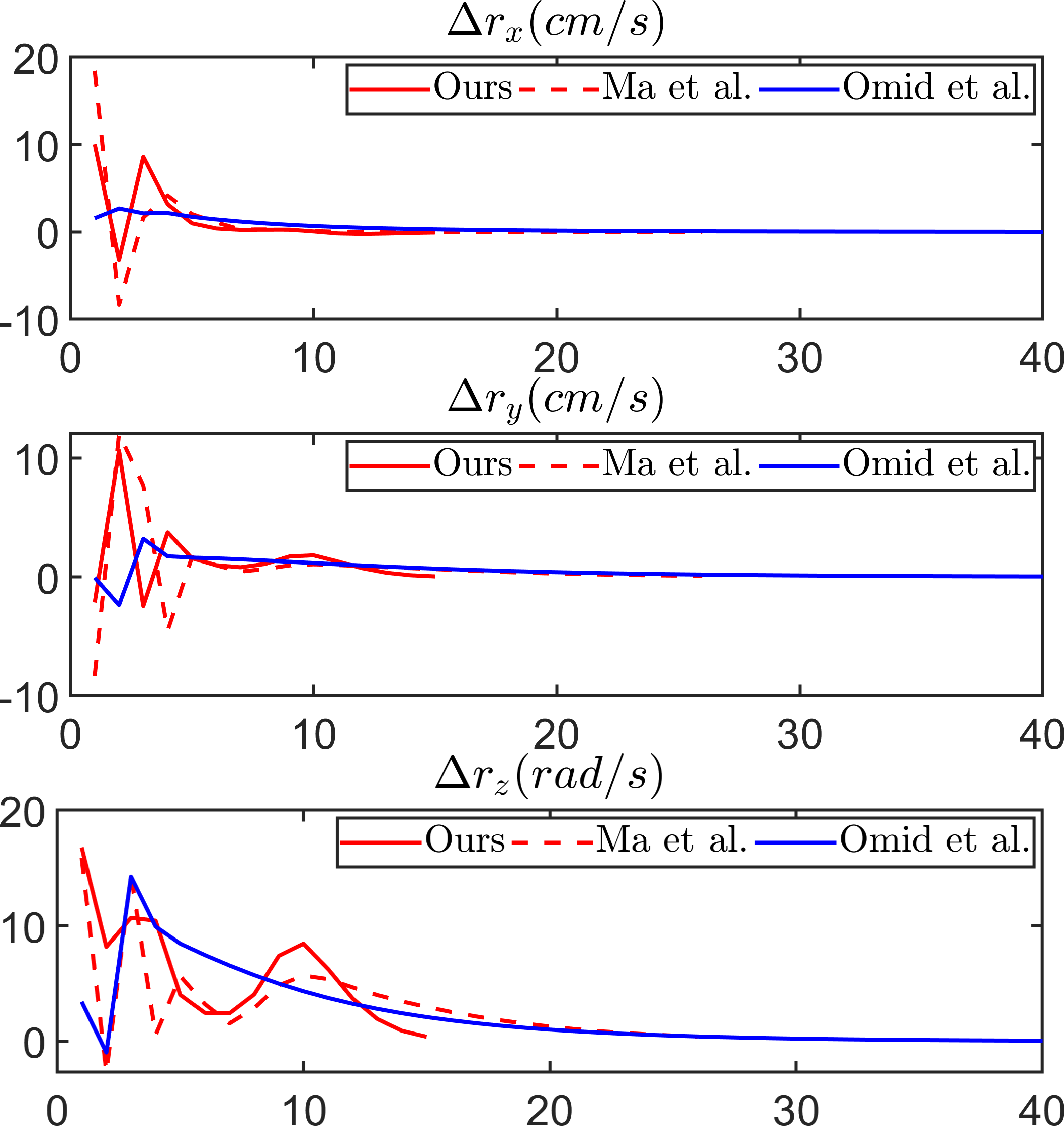}
\caption{
Profiles of the velocity command $\Delta \mathbf{r} $ error $T_1$ among Ours, Ma et al. \cite{ma2022active}, and Omid et al. \cite{aghajanzadeh2022optimal}.
The abscissa is the step size.
}
\label{fig2}
\end{figure}

Fig. \ref{fig3} gives the deformation trajectories of the elastic rod among Ours, Ma et al. \cite{ma2022active}, and Omid et al. \cite{aghajanzadeh2022optimal}.
From this viewpoint, the proposed manipulation framework shows the fastest manipulation performance, which follows \cite{ma2022active}.
The slowest convergence is presented in \cite{aghajanzadeh2022optimal}, but it has the smoothest profiles.

Fig. \ref{fig1} presents the profiles of the deformation error $T_1$ of the utilize three manipulation frameworks.
Ours has the smallest convergence steps for the numerical results, while \cite{aghajanzadeh2022optimal} presents the smoothest decrease trajectory.
Meanwhile, ours has the fastest convergence speed near the neighborhood of the origin, compared to the other two approaches.

Fig. \ref{fig2} displays the profiles of the velocity command $\Delta \mathbf{r}_k$ among three frameworks.
$\Delta r_x$ and $\Delta r_y$ are the linear speed of the effector, and $\Delta r_z$ is the rotation speed of the effector around $z$-axis.
From these results, it can be seen that ours has the best control effect with the fastest convergence speed and has excellent adaptability to various conditions in shape deformation.

\appendices
\ifCLASSOPTIONcaptionsoff
  \newpage
\fi

\bibliography{biblio.bib}
\bibliographystyle{IEEEtran}
\end{document}